\documentclass{article}

\usepackage[utf8]{inputenc}
\usepackage[TS1,T1]{fontenc}
\usepackage{lmodern}

\DeclareUnicodeCharacter{1EB9}{\d{e}}
\DeclareUnicodeCharacter{1ECD}{\d{o}}

\newcommand{\s}{{\fontencoding{TS1}\selectfont\char115}} 

\usepackage{hyperref}

\usepackage[frozencache,cachedir=.]{minted}

\usepackage{tikz}

\usepackage[french]{babel}

\usepackage{csquotes}
\usepackage[backend=biber, sorting=nyt, mincitenames = 1, maxcitenames = 1, maxbibnames = 100, style=authoryear-comp]{biblatex}

\DefineBibliographyExtras{french}{\renewcommand*\mkbibnamefamily[1]{#1}}

\title{La production de corpus d'occitan médiéval et prémoderne: problèmes et perspectives de travail\thanks{%
    Les données et scripts utilisés pour cet article sont disponibles sur
    Zenodo, \url{https://zenodo.org/record/2605497}, \textsc{doi}: \texttt{10.5281/zenodo.2605497}. 
    Les développements ultérieurs du corpus seront consultables sur \textit{Github}, \url{https://github.com/lemma-oc/BOMM}.
}}

\author{Jean-Baptiste Camps\\ Centre Jean-Mabillon \\ École nationale des chartes | Université PSL  \and Gilles Guilhem Couffignal\\ Sorbonne Université | EA 4509, STIH }

\date{Actes du XII\ieme{} Congrès de l'Association internationale d'études occitanes\\ Albi, 2017}

\newcommand{\graph}[1]{‹#1›}

\begin{document}
	
	\maketitle
	
Le chercheur en langue et littérature occitanes est déjà bien bien pourvu en matière de corpus textuels électroniques, de façon même surprenante si l'on considère le statut social de l'occitan dans la société française actuelle. Le médiéviste occitaniste, notamment, a depuis longtemps à sa disposition la précieuse COM, fruit d'un travail pionnier mené en son temps par le regretté Peter Ricketts \parencites{ricketts_concordance_2001}{ricketts_concordance_2005}. D'autres projets de corpus \parencites{field_corpus_2016}{scrivner_old_2016} ou d'édition électronique \parencites{lodge_les_2006}{carrasco_2014}
touchant à la période médiévale ont également fleuri ces dernières années.

Pour les études modernes, 
les projets sont également nombreux et anciens. On peut les faire remonter, dans une perspective computationnelle, aux tables dialectométriques de  \cite{seguy_1971}\footnote{Pour les corpus \textit{stricto sensu}, 
voir dans ce même volume les projets en cours (Batelòc, CoRLiG); on notera également des projets de bibliothèque virtuelle \parencite{nepote_bibliotheca_2006}.}.
Les projets ne manquent pas, ni même les sources déjà traitées et donc exploitables. Devant la disparité des techniques employées, nous voudrions montrer ici comment une équipe restreinte de deux ou trois chercheurs peut concrètement travailler à partir de corpus textuels électroniques. Deux points nous intéresseront particulièrement, l'acquisition et le traitement des données. 

Pour l'acquisition des données, nous développerons la question de la reconnaissance optique des caractères (\textit{optical character recognition} ou	\textsc{ocr}) et de la reconnaissance des écritures manuscrites (\textit{handwritten text recognition} ou \textsc{htr}), qui est la technique permettant aujourd'hui à n'importe quel chercheur muni d'un texte numérisé et d'un terminal informatique de créer rapidement son propre corpus, moyennant quelques manipulations informatiques et opérations de transcription que nous détaillerons.

En termes de traitement des données, nous retiendrons l'exemple de la lemmatisation, encore peu souvent effectuée sur les corpus occitans, et pivot d'un grand nombre d'exploitations possibles. Ce sera également l'occasion de montrer toute l'importance du partage des données et de formats interopérables entre les différents projets.

\section{L'acquisition de corpus: transcription et reconnaissance optique de caractères}

Le développement d'outils ouverts de reconnaissance optique des caractères, fondés sur des méthodes d'apprentissage profonds permet de les adapter à la situation d'un imprimé ancien, d'un incunable, voire d'un manuscrit, avec des résultats exploitables  \parencites{breuel_ocropy:_2014}{springmann_ocrocis:_2015}, ce qui autorise la constitution rapide de \textit{corpora} textuels. 

La mise en œuvre d'une série de phases d'entraînement, correction, et nouvel entraînement permet à un chercheur de faire progresser la qualité des résultats, en accumulant rapidement des données textuelles brutes. Des outils libres et performants sont désormais à la portée des chercheurs en SHS (Ocropy/CLSTM, Tesseract 4) et permettent d'atteindre des niveaux d'erreur caractère (\textsc{cer} ou \textit{character error rate}) qui peuvent relativement aisément descendre sous les 2\% pour des imprimés anciens, et sous les 10\% pour des manuscrits médiévaux. La seule limite est que si des modèles existent déjà pour les documents contemporains, il faut le plus souvent les développer soi-même pour les états de langue anciens, voire les adapter à chaque main en ce qui concerne les manuscrits. 

\subsection{Présentation générale}

\begin{figure}
    \centering
    \resizebox{0.8\textwidth}{!}{%
	\begin{tikzpicture}
	\node[draw, rectangle, rounded corners=3pt] (P) at (0,6)
	{Prétraitement des images};%
	\node[draw, rectangle, rounded corners=3pt] (A) at (0,5)
	{Analyse mise en page};%
	\node[draw, rectangle, rounded corners=3pt] (GT) at (5,5)
	{Vérité de terrain};
	\node[draw, rectangle, rounded corners=3pt] (T) at (7,3)
	{Entraînement};
	\node[draw, rectangle, rounded corners=3pt] (TT) at (3,3)
	{Test};
	\node[draw, rectangle, rounded corners=3pt] (O) at (3,1.5)
	{Sortie};
	
	\draw[->,>=latex] (P) -- (A) ;
	\draw[->,>=latex] (A) -- (GT) node[midway,sloped,above] {transcription}; 
	\draw[->,>=latex] (GT) -- (T) ; 
	\draw[->,>=latex] (T) -- (TT) ;
	\draw[->,>=latex] (TT) -- (GT) node[midway,sloped,above] {relecture};
	\draw[->,>=latex] (TT) -- (O) ;
	\end{tikzpicture}
}
    \caption{Vue simplifiée des étapes de l'entraînement d'un outil d'\textsc{ocr}/\textsc{htr}}
    \label{fig:OCRopySteps}
\end{figure}
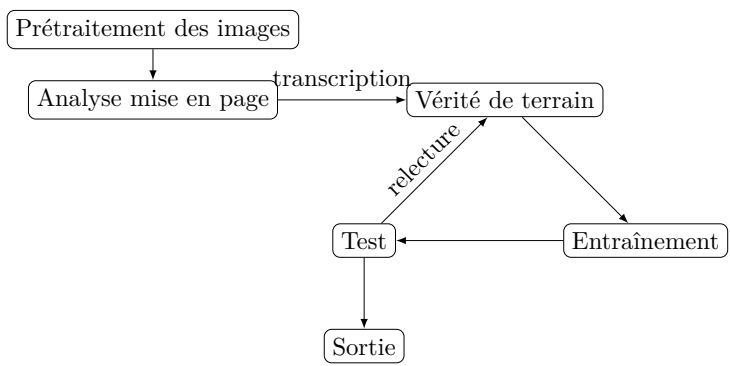

Le processus de reconnaissance d'un texte (\textsc{ocr} ou \textsc{htr}) se décompose généralement en plusieurs étapes (fig.~\ref{fig:OCRopySteps}),
\begin{enumerate}
	\item le traitement des images: recadrage, redressement, suppression du bruit, et binarisation (en différenciant le texte de l'arrière-plan de l'image);
	\item l'identification de la mise en page: zones de texte, colonnes, lignes, etc.;
	\item la reconnaissance proprement dite du texte (i.e., la ``transcription automatique''), à partir d'un modèle préexistant, ou bien en alternant phase d'entraînements d'un modèle et phase d'accroissement des données (reconnaissance de nouvelles lignes, correction,…);
	\item d'éventuels post-traitements, visant à améliorer les résultats, en se fondant sur des modèles linguistiques (lexiques, …) ou en traitant en série des erreurs fréquentes.
\end{enumerate}

La première étape peut être réalisée avec un outil dédié au traitement des images, avec ou sans interface graphique. Certains de ces outils sont d'un maniement très simple pour le néophyte, tels que \texttt{ScanTailor} \parencite{artsimovich_scantailor_2016}. 
Cette étape n'est pas du tout anodine, et la qualité des résultats que l'on obtiendra en \textsc{ocr}/\textsc{htr} est directement liée à la qualité des images traitées. Il est donc primordial non seulement d'obtenir des fichiers de bonnes qualité (300 ou 600 points par pouces), mais encore des les rendre les plus lisibles possible: redressement de l'image; suppression de sources de bruit (automatiquement ou par intervention humaine pour supprimer la décoration ou les partitions musicales, comme pour l'\textsc{ocr} des \textit{Psaumes}  d'Arnaud de Salette); binarisation, pour accentuer les contrastes et isoler le texte du bruit d'arrière-plan ou des lettres du verso (quand l'encre corrosive les fait apparaître en transparence).

Enfin, le texte est segmenté par colonne et par ligne, unité de base du traitement de reconnaissance.
L'identification de la mise en page, et particulièrement des colonnes et des lignes, peut être plus délicate, notamment à cause des déformations possibles des images, des irrégularités que présentent les manuscrits, ou de la présence d'éléments tels que les lettrines, qui occupent plusieurs lignes de réglure.
Pour cette phase d'analyse de la mise en page, une nouvelle génération d'algorithmes, fondés sur de l'apprentissage machine, est en train d'émerger. 

Pour ce qui est de la reconnaissance du texte proprement dite, les outils existants peuvent être subdivisées en deux catégories, selon qu'ils aient recours à des méthodes segmentées ou non segmentées \parencites{breuel2013high}[11-18]{ul-hasan_generic_2016}.
Les méthodes segmentées visent tout d'abord à décomposer la ligne en une série de caractères, que l'on tentera ensuite d'identifier en les comparent individuellement à des formes enregistrées pour chaque caractère.
Les méthodes non segmentées se fondent sur une analyse de la ligne, sans segmentation préalable, permettant ainsi une meilleure prise en compte du contexte. Elles peuvent, par exemple, faire défiler une fenêtre de largeur fixe sur la ligne en combinaison avec des modèles probabilistes (modèle de Markov caché). Les méthodes non segmentées actuellement les plus prometteuses ont recours à l'apprentissage de séquence, grâce à des réseaux de neurones, et tout particulièrement des réseaux récurrents  à mémoire court et long terme (\textit{long short-term memory} ou LSTM), uni- ou bidirectionnels, en combinaison avec un algorithme de de
classification temporelle connexionniste (CTC) \parencites{breuel2013high}{ul-hasan_generic_2016}.

Nous avons ainsi choisi d'utiliser un outil libre implémentant ce dernier type de méthode, \texttt{OCRopy} \parencite{breuel_ocropy:_2014}, de pair avec son implémentation en C, \texttt{CLSTM}.

\subsection{Un outil libre: OCRopy}	
	
\begin{figure}
    \centering
    \includegraphics[width=0.9\textwidth]{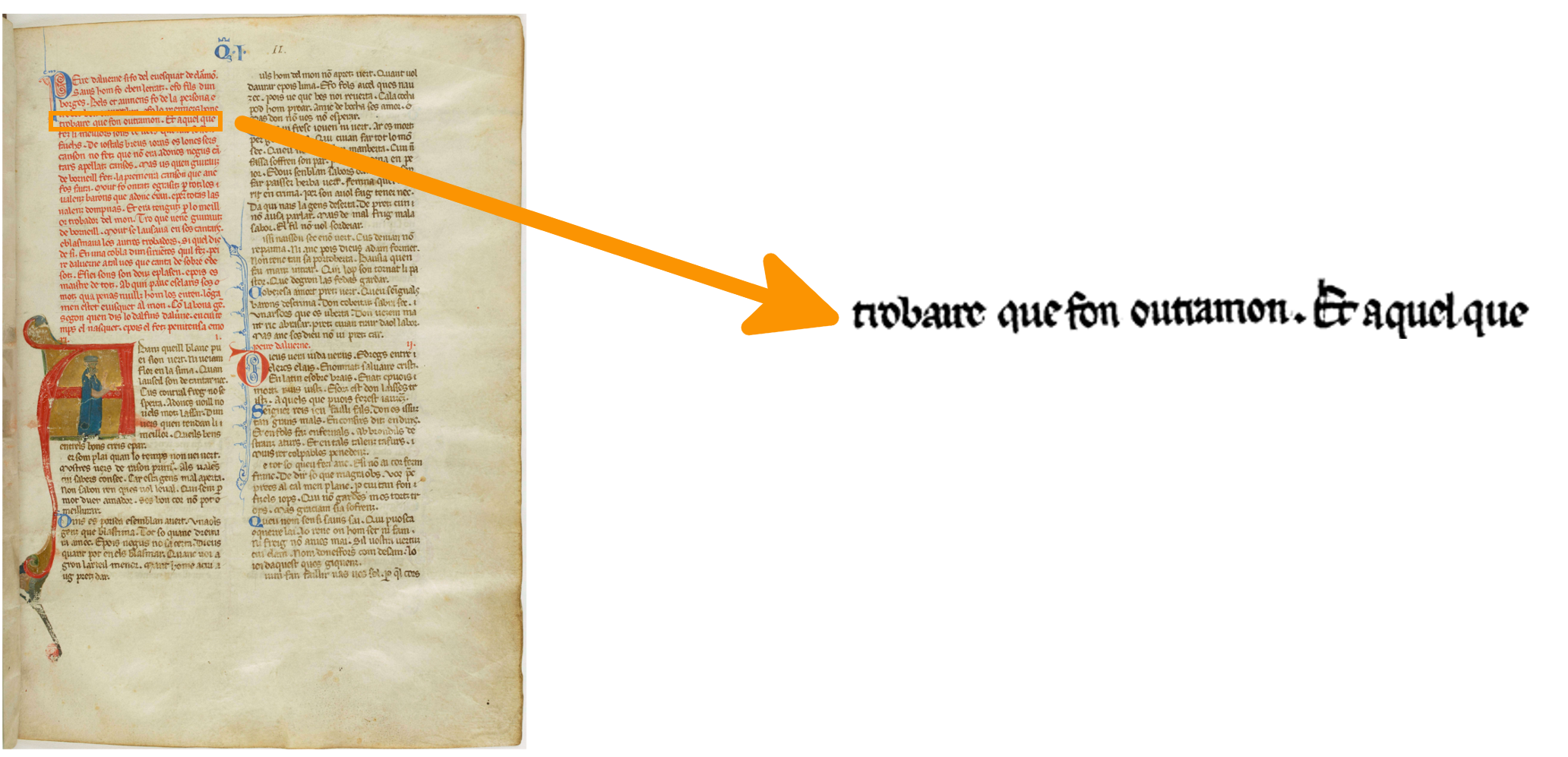}
    \caption{Extractions des lignes binarisées d'un manuscrit (chans. I, BnF fr.~854)}
    \label{fig:pretraitementChans}
\end{figure}

\begin{figure}
    \centering
    \includegraphics[width=0.45\textwidth]{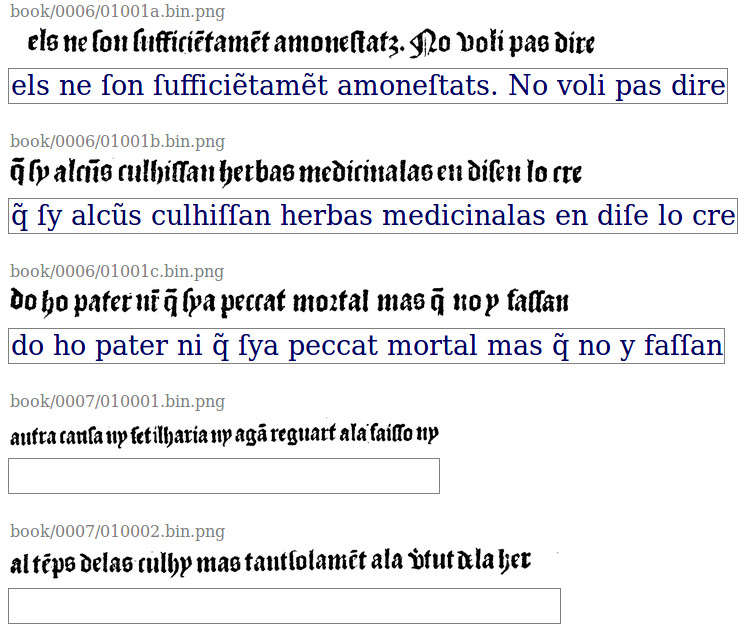}
    \includegraphics[width=0.45\textwidth]{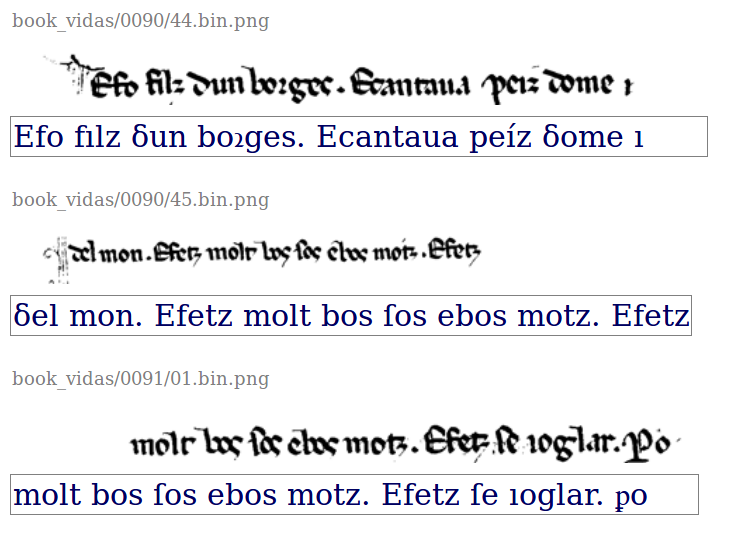}
    \caption{Production de données de vérité de terrain pour un imprimé ancien (gauche, \textit{Doctrinal}, 1504) et un ms. (droite, transcr. allographétique du chans. I)}
    \label{fig:transcr}
\end{figure}

\texttt{OCRopy} est un outil \textit{open source}, avec lequel il est aisé d'entraîner et réutiliser des modèles de reconnaissance des caractères ou des écritures.
L'entraînement de modèles par apprentissage machine repose sur des données initiales transcrites par l'humain (la vérité de terrain, en anglais, \emph{ground truth}). Concrètement, après avoir pré-traité le document numérique de façon à générer un fichier image par ligne de texte, 
le philologue est tout d'abord invité à transcrire une sélection pertinente (d'un point de vue graphique ou typographique) du texte (fig.~\ref{fig:pretraitementChans} et \ref{fig:transcr}).
			
Cette transcription se fait selon les règles choisies par l'opérateur. Elle peut être de type ``diplomatique''. Nous préférons les termes d'allographétique et graphématique, suivant la terminologie proposée par \cite{stutzmann_paleographie_2011}. Dans le premier cas chaque allographe est encodé par son propre caractère, tandis que dans le second cas, seul le graphème est représenté par un caractère informatique. Un exemple courant: tous les allographes de \graph{s} peuvent soit être rendus par un caractère qu'on leur consacre (\s,…), soit par un seul caractère (s).
	
Tous les signes nécessaires à cette transcription sont déclarés dans un ficher dédié (ici, \texttt{chars.py}). Ainsi, les seules limites à une approche allographétique seront celles de la modélisation réalisée par le chercheur (quels sont les allographes repérables?
Combien de variantes en recense-t-on?) et de l'existence d'une convention pour la représentation de ces signes (idéalement au sein d'Unicode, ou bien d'une initiative collective comme la \textit{Medieval Unicode Font Initiative}, \cite{haugen_medieval_2014}).
Quoiqu'on ait choisi, et si l'on ne veut obtenir plus aisément des résultats, il est préférable de transcrire ce qui figure effectivement sur la source: c'est-à-dire notamment les abréviations (et non leur résolution), la ponctuation et la segmentation originale\footnote{Demander à l'intelligence artificielle de résoudre d'elle-même les abréviations n'est pas impensable, mais bien plus difficile, et demande beaucoup plus de données  \parencite[voir][]{bluche2017preparatory}. Les scripts de résolution d'abréviation, en revanche, sont beaucoup plus facilement appliqués en post-traitement, ce qui permet, en outre, de conserver dans un même fichier l'état observé et l'état interprété.}. 

L'étape suivante consiste en la transcription des lignes (ou la correction de lignes prédites par le modèle précédant), permettant de fournir des données d'entraînement en volume suffisant (fig.~\ref{fig:transcr}).
			
\begin{figure}
    \centering
    \includegraphics[width=0.8\textwidth]{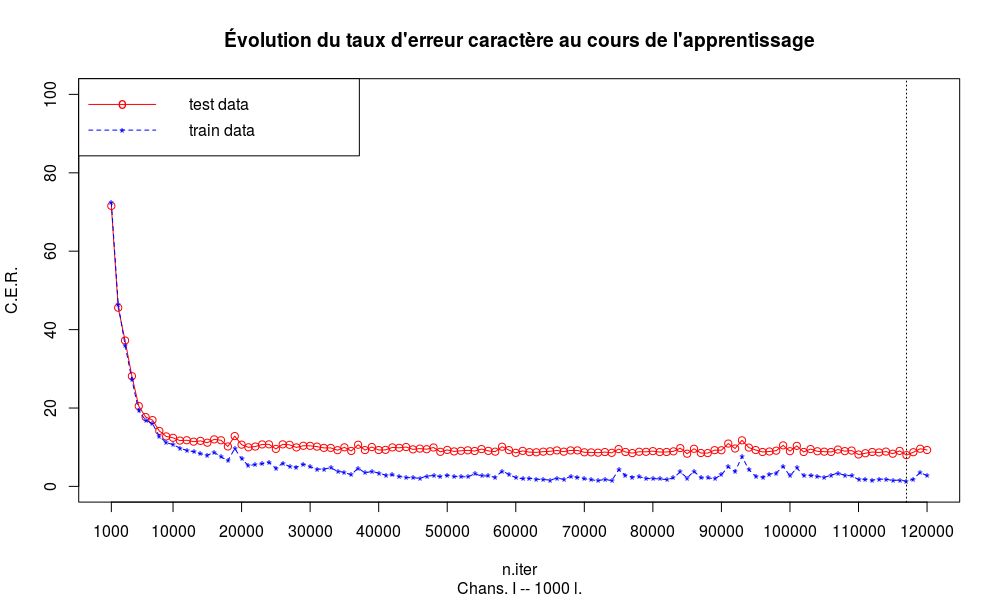}
    \caption{Courbes d'apprentissage du second entraînement sur les données du chansonnier \textit{I}, avec 1~000 lignes transcrites. Le meilleur modèle, à la fois sur les données de test et d'entraînement, apparaît à l'itération 117~000}
    \label{fig:trainI}
\end{figure}

Une fois ces données initiales apprêtées par un humain, la machine prend le relais et l'entraînement peut ainsi commencer. Lors de chaque itération de celui-ci, la machine tente d'aligner le texte de la vérité de terrain avec l'image (algorithme \textsc{ctc}) et de prédire ce texte. Le réseau de neurones est ensuite mis à jour en fonction des résultats obtenus.
Cet apprentissage se poursuit alors durant plusieurs dizaines ou centaines de milliers d'itérations, jusqu'à obtenir un résultat satisfaisant. La quantité de données disponibles est très importante si on veut retarder ce que l'on nomme l'\textit{overtraining}, c'est-à-dire le moment où le réseau \og{}apprend par cœur\fg{} les données d'entraînement, et commence à régresser sur les données de test (cf. fig.~\ref{fig:trainI};  après l'itération 120~000, l'erreur -- non portée sur le graphique -- sur les données de test recommence à augmenter).

À partir de ce moment, se met en place un va-et-vient entre l'humain et la machine. Les différents modèles enregistrés par le programme sont évalués. C'est-à-dire qu'ils sont appliqués, en aveugle, sur des données sélectionnées de façon aléatoire et déjà transcrites mais non vues lors de l'entraînement, de façon à ce que l'on puisse compter les erreurs. 

Le meilleur modèle est alors retenu et appliqué à une autre sélection du texte. L'opérateur peut alors, très rapidement, corriger cette sélection et ainsi créer plus de données vérifiées qui permettront une nouvelle série d'entraînements.
Lorsque le dernier modèle entraîné est jugé suffisamment performant, on l'applique à l'ensemble de l'ouvrage et on récupère l'ensemble des données. Dans le cas d'ouvrages particulièrement lisibles et pour lesquels on peut prétendre arriver à plus de 99\%{} de réussite, il est préférable de se satisfaire du texte créé par l'OCR. Il y a de grandes chances pour que certaines menues erreurs du transcripteur humain aient été rectifiées par la machine. En revanche, dans le cas d'ouvrages plus complexes ou dégradés, il est préférable de recourir aux données machine pour compléter celles de l'humain.

Pour un imprimé ancien, on peut raisonnablement atteindre des taux d'erreur de l'ordre de 1\%\footnote{Durant nos essais sur des imprimés occitans du \textsc{xvi}\ieme{} siècle, seul le modèle des \emph{Salmes} de Salette (1587) est resté à un taux d'échec supérieur à 2 (2,091\%), dû sûrement à la mise en page complexe de l'ouvrage. Nous sommes parvenus, pour la traduction de Gerson (1556) à 0,797\% d'échec.}, ce qui permet immédiatement l'application de méthodes de recherche reposant sur la quantité de données textuelles non vérifiées.%
Dans le cas du manuscrit, une relecture reste impérative. Néanmoins, le temps nécessaire, par rapport à une transcription brute, est très fortement diminué dès lors que l'on brise la barrière des 10\% d'erreur, et à plus forte raison encore, de 5\%. Si le premier seuil n'est pas hors de portée avec quelques centaines seulement de lignes d'entraînement, le second demande une quantité bien plus importante de données.

\begin{figure}
    \centering
    \includegraphics[width=0.5\textwidth]{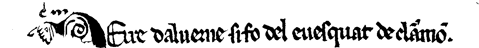}
		\includegraphics[width=0.5\textwidth]{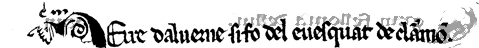}		
		\includegraphics[width=0.5\textwidth]{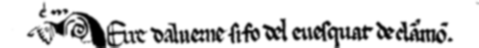}
		\includegraphics[width=0.5\textwidth]{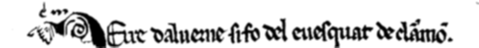}
		\includegraphics[width=0.5\textwidth]{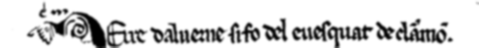}
		\includegraphics[width=0.5\textwidth]{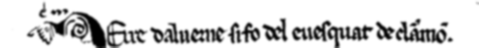}
		\includegraphics[width=0.5\textwidth]{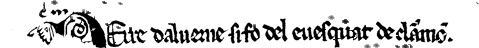}
		\includegraphics[width=0.5\textwidth]{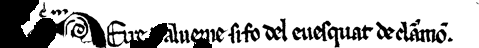}
		\includegraphics[width=0.5\textwidth]{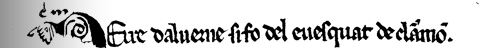}
		\includegraphics[width=0.5\textwidth]{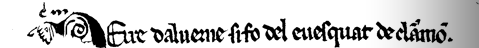}
    \caption{Déformations appliquées, avec \texttt{DocCreator}, aux images de vérité de terrain du chansonnier \textit{I}. De haut en bas: ligne originale, simulation de l'apparition du verso en transparence, quatre formes de flou, dégradation des caractères, simulation de trous et simulation de l'ombre de la reliure à gauche et à droite. Des simulations de déformation 3-D (page froissée…) sont possibles également.}
    \label{fig:syntheticGT}
\end{figure}

Comme la constitution des données de terrain (transcription, relecture) est ce qui est le plus coûteux en temps et compétences, diverses solutions peuvent être employées pour en augmenter artificiellement le nombre, soit en appliquant diverses formes de transformations aux images des lignes, soit en créant des pseudo-pages de manuscrits à partir de l'extraction de la forme des lettres. Ce travail peut être réalisé en quelques minutes grâce à un logiciel tel que \texttt{DocCreator} \parencites{journet2017doccreator}{labri_doccreator:_2018}. Si l'on retient la première solution, on pourra appliquer une variété de déformations aux images originelles, y compris des déformations tridimensionnelles (fig.~\ref{fig:syntheticGT}).

Dans le cas du chansonnier \textit{I}, 1000 lignes d'entraînement suffisent à obtenir un premier modèle tout à fait exploitable, tandis que l'ajout de lignes artificielles permet une amélioration supplémentaire de presque 2\% de \textsc{cer} (table~\ref{tab:resultsHTR}).

\begin{table}
    \centering
\begin{small}
\begin{tabular}{c|ll}
		& Modèle 1 & Modèle 2 \\ \hline \hline
		vérité de terrain (l.) & 1000 & 1000\\
		vér. de terr. artificielle (l.) & 0 & 9000 \\
		test (l.) & 100 & 100 \\
		n. itér. &  117k & 295k\\
		CER entr.. & 1.3 & 3.5* \\
		\textbf{CER test} & \textbf{8} & \textbf{6.1} \\
		sans espaces & 7.4 & 5.7 
	\end{tabular}	
\end{small}
    \caption{Résultats des meilleurs modèles pour les entraînements sur le chans. \textit{I}.\newline *. données artificielles exclues dans le calcul du \textsc{cer} entr.}
    \label{tab:resultsHTR}
\end{table}

Que reste-t-il alors comme erreurs?
Les confusions les plus courantes que nous ayons pu observer pour les manuscrits romans ne sont pas surprenantes (table~\ref{tab:confusions} pour le chansonnier \textit{I}). Les erreurs les plus fréquentes concernent ainsi les caractères ``combinatoires'' --~accents et tildes variés~--, ainsi que l'espacement. On sait que l'opposition présence/absence d'une espace, propre à l'imprimé, ne correspond pas à la réalité du manuscrit et pose des difficultés au transcripteur qui voudrait reproduire la pratique médiévale. Ces difficultés réapparaissent aussi pour un réseau de neurones.
D'autres confusions peuvent s'expliquer par des proximités de forme, entre  \textit{e} et \textit{o} par exemple.
		
\begin{table}
    \centering 
\begin{small}
    \begin{tabular}{ccc} 
freq. & pred. & GT \\ \hline \hline
21&\_&[diacr.]\\ 
12& [espace] &\_\\ 
9&\_& [espace] \\ 
6&[diacr.]&\_\\ 
5&o&e\\ 
4&\_&ı\\ 
4&e&o
	\end{tabular}
\end{small}
    \caption{Matrice des confusions les plus fréquentes obtenues par le meilleur modèle sur les données de test (chansonnier \textit{I}). La colonne de gauche donne ce qui a été prédit par le modèle, et celle de droite la vérité de terrain. Le symbole `\_' signale une absence de caractère.  }
    \label{tab:confusions}
\end{table}

\section{Structuration de documents textuels}
	
Une fois les données acquises, se pose la question de leur représentation par un modèle, qui est nécessairement une simplification de la réalité, ou pour mieux dire une sélection de faits observables sur le document source, que nous retenons parce que nous voulons les analyser ou parce qu'ils sont d'un intérêt suffisamment général pour justifier le surcroît de travail \parencites{sperberg2009teach}[cxcv-cc]{camps2016chanson}.
	
Les sources présentent ainsi une grande diversité de faits, de différents ordres:
	\begin{itemize}
		\item matériel (cahiers, feuillets, dégâts touchant au support,…) 
		et relatif à la mise en texte et en page (colonnes, lignes,…);
		\item logique (paragraphes, strophes, vers…);
		\item graphique (segmentation, allographie, ponctuation,…);
		\item métrique, linguistique, etc.
	\end{itemize}
		
La première étape de la structuration des données est ainsi la conception d'un modèle abstrait, exprimant les concepts retenus et leurs relations, qui peut ensuite être retranscrit par un formalisme technique donné. Dans ce contexte, les recommandations de la \textit{Text encoding initiative} 
\parencite{tei_consortium_tei_2015} 
fournissent un cadre dans lequel peut se placer l'implémentation du modèle.

La conformité à la TEI n'est toutefois pas suffisante en soi: il reste au chercheur à concevoir son modèle comme un sous-ensemble plus spécifique de celle-ci, et documenté selon l'usage qu'il en fait.
Pour que ce modèle soit utile, il ne doit pas simplement viser à la satisfaction intellectuelle de son créateur, mais doit demeurer manipulable,
permettre l'exploitation scientifique, la publication et la réutilisation des données, ainsi que leur pérennisation et, idéalement, l'échange aisé entre différents projets.
D'un point de vue plus général, il est souhaitable de se conformer aux principes de la science ouverte et des \og{}\textsc{fair} data\fg{}, données \og{}trouvables, accessibles, interopérables et réutilisables\fg{} \parencite{wilkinson2016fair}.

Une fois le modèle de données établi, il convient de structurer les documents d'une manière conforme à celui-ci (ce que facilite, en cas d'utilisation de technologies \textsc{xml}, l'utilisation d'un schéma dans un langage tel que \texttt{RelaxNG}). Cette étape de structuration peut être sujette à différentes automatisations. Il est ainsi possible \textit{a minima} de recourir à des scripts dans un langage de programmation, tel que \texttt{Python} ou \texttt{xslt} par exemple, pour une première phase d'encodage sommaire en tirant profit des informations fournies par l'\textsc{ocr} ou déterminables par des motifs et expressions régulières (segmentation en feuillets, lignes, strophes, vers, mots, …).

Si un encodage automatique de ce type ne permet pas d'aller très loin, des pistes existent pour permettre un encodage fin de documents par la machine, grâce à des méthodes d'intelligence artificielle. Des outils comme \texttt{Grobid} \parencite{romary2015grobid}
permettent déjà de traiter de cette manière des documents très structurés, tels que des publications scientifiques ou des dictionnaires, à partir d'un échantillon encodé de données d'entraînement.

Dans le cas le plus général, pour ce qui est des éditions de manuscrits ou imprimés anciens, une phase d'intervention humaine sur les fichiers s'avère nécessaire, quoique coûteuse et chronophage. Il reste néanmoins possible d'enrichir systématiquement les données transcrites par des méthodes computationnelles, qu'il s'agisse de la reconnaissance des entités nommées ou de la lemmatisation.

\section{Enrichir les données transcrites: l'exemple de la lemmatisation}
	
Une infinité d'enrichissements peuvent être accomplis sur les données issues de la transcription, certains plus courants que d'autres: reconnaissance des entités nommées, annotation du système graphique, des morphèmes, du mètre, des figures de style, etc. Parmi ces enrichissements, nous nous attarderons sur un cas très fréquent, et utile à de nombreuses formes d'exploitation des données, à savoir la lemmatisation et l'annotation morpho-syntaxique.
	
Cet enrichissement des données textuelles par des annotations d'ordre linguistique peut également connaître certaines automatisations. Si ces procédés sont bien connus des linguistes et déjà bien fonctionnels pour les langues modernes, ils posent encore des difficultés pour des langues comme l'occitan médiéval, dues en partie à l'absence de corpus lemmatisés de taille conséquente. 
	
\subsection{Principe général}
	
Les lemmatiseurs fondés sur des lexiques de forme sont d'un emploi courant  pour les langues contemporaines 
\parencite[par ex.][]{schmid_treetagger:_1994}, mais ceux-ci s'avèrent inadaptés pour des états anciens de langue présentant une forte variation graphique et de nombreuses formes en hapax, et rarement capables de désambiguïser des formes homographes. 
La très grande variation graphique réduit sensiblement la pertinence des lemmatiseurs fondés sur des lexiques de formes, sachant que des substitutions graphiques nouvelles peuvent apparaître dans chaque nouveau jeu de données soumis au lemmatiseur.
	
En revanche, la nouvelle génération de lemmatiseurs, 
qui tire profit des avancées de l'apprentissage machine, est plus prometteuse pour les langues médiévales ou riches en variation. Ceux-ci se fondent souvent sur des corpus annotés --~soit des occurrences prises en contexte~--, dont ils cherchent à modéliser l'espace sémantique et la variation graphique \parencites{muller2015joint}{kestemont_lemmatization_2016}. Ces lemmatiseurs sont ainsi capables de réaliser la tâche  centrale à la lemmatisation d'états anciens de langue: l'identification correcte du lemme d'une forme inconnue (i.e. jamais rencontrée durant l'entraînement). Ils peuvent aussi décider correctement du lemme d'une forme qui possède plusieurs homographes. 
	
\subsection{Collecte et alignement des données}
	
Pour entraîner ces lemmatiseurs, il faut donc disposer de corpora de données annotées, en quantité et qualité suffisantes. Ce besoin se heurte d'emblée à une première difficulté, celle de l'interopérabilité des jeux de données, en termes de formats, mais aussi en termes de référentiels, particulièrement les listes de lemmes de référence et les jeux d'étiquette. Des barrières d'usage ou juridiques surgissent également rapidement, car l'accès aux données annotées ou la possibilité de les réutiliser n'est pas systématique. Tandis que certaines données peuvent être emprisonnées dans des systèmes propriétaires, d'autres ne sont simplement pas accessibles pour des raisons techniques ou parce que la démarche de les doter d'un statut et d'une licence claire et explicite n'a pas été accomplie.
Sur un plan pratique, se pose en outre la question de la quantité de données nécessaires. Il est en effet délicat de parvenir à des résultats dotés d'un peu de généralité sans un corpus dont le nombre d'occurrences n'est pas un multiple de $10^5$, ou, de préférence, $10^6$ -- à titre de comparaison, la totalité de la \textsc{com1} \parencite{ricketts_concordance_2001} contient environ $7,3 \times 10^5$ occurrences.
Dans les lignes qui suivent, nous décrirons comment nous avons constitué un corpus d'entraînement primaire (annoté), ainsi qu'un corpus secondaire (non annoté) pouvant être utilisé plus marginalement pour ajuster les représentations sémantiques des mots. 
	
\begin{figure}
	    \centering
	    \begin{tikzpicture}
		\node[draw, rectangle, rounded corners=3pt, text width=2.5cm, text centered] (A) at (0,2) {Récupération des sources};
		\node[draw, rectangle, rounded corners=3pt, text width=2.5cm, text centered] (B) at (4,2) {Alignement (TEI + DOM)};
		\node[draw, rectangle, rounded corners=3pt, text width=2.5cm, text centered] (C) at (8,2) {Relectures et corrections};
		\node[draw, rectangle, rounded corners=3pt, text width=2.5cm, text centered] (D) at (2,0) {Entraînement du lemmatiseur};
		\node[draw, rectangle, rounded corners=3pt, text width=2.5cm, text centered] (E) at (6,0) {Annotation de nouvelles données};
		
		\draw[->,>=latex] (A) -- (B);
		\draw[->,>=latex] (B) -- (C);
		\draw[->,>=latex] (C) -- (D);
		\draw[->,>=latex] (D) -- (E);
		\draw[->,>=latex] (E) to[bend right] (C);
		\end{tikzpicture}
	    \caption{Étapes de l'entraînement d'un lemmatiseur}
	    \label{fig:etapesLemmat}
	\end{figure}
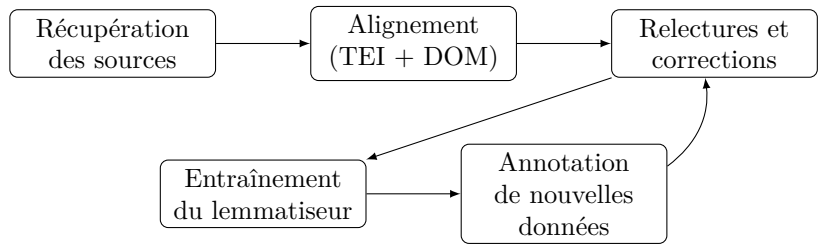

Dans le cadre de notre démarche d'entraînement d'un lemmatiseur pour l'occitan médiéval et prémoderne, la première étape à été de tenter de collecter des données d'entraînement, héritées de divers projets.
Nous avons ainsi utilisé les données fournies, sous licence libre, 
par l'édition des \textit{Comptes des consuls de Montferrand} \parencite{lodge_les_2006}, ainsi que deux corpora qui nous ont été fournis par leurs auteurs \parencites{field_corpus_2016}{scrivner_old_2016}\footnote{%
    À la date d'écriture de cet article, nous n'avons pas encore eu recours au corpus fourni par \cite{field_corpus_2016}, n'ayant encore pu compléter de manière systématique les informations qui y figuraient et les aligner sur le référentiel commun utilisé.
}.
Il nous a ensuite fallu procéder selon les étapes présentées en fig.~\ref{fig:etapesLemmat}.

Comme ces données étaient dans des formats différents, l'étape suivante est de les aligner sur un format commun, avec un même jeu de lemmes de référence.
	
Les données des \textit{Comptes…} se présentaient déjà sous un format \textsc{tei}, en partie seulement segmenté et lemmatisé, via des liens avec les entrées du glossaire (fig.~\ref{fig:MontfOrig}). Nous avons donc procédé, par un script \texttt{xslt}, à une segmentation complète, et avons aligné les entrées du glossaire avec le jeu de lemmes de référence que nous avons retenu, à savoir celui du \textit{Dictionnaire de l'occitan médiéval} \parencite{stempel_dictionnaire_1996}, comme présenté en fig.~\ref{fig:alignLemmesMontf}. Pour ce faire, un script a interrogé systématiquement les entrées du \textsc{dom} à partir de l'entrée du glossaire, et proposé une ou plusieurs correspondances possibles, qui ont été vérifiées par l'humain et intégrées aux données (ligne 3 sur la figure). Dans les cas où aucun lemme du \textsc{dom} ne correspondait à l'occurrence, nous avons créé et documenté de nouveaux lemmes, en tentant de suivre les mêmes principes de création que le \textsc{dom}.

\begin{figure}
\begin{minted}[linenos]{xml}
<item xml:id="CC6.278">Item 
    IIII s. qu <w lemmaRef="#gloss_a116_11">ac</w> 
    D. Chapus <w lemmaRef="#gloss_d57">dos</w> II jorns 
    <w lemmaRef="#gloss_d40">desus</w> 
    <w lemmaRef="#gloss_d49_9">dit</w> 
    <w lemmaRef="#gloss_q11">que</w> 
    <w lemmaRef="#gloss_a47_4">anet</w> 
    <w lemmaRef="#gloss_e9">en</w> Garsias 
    <w lemmaRef="#gloss_p30">per</w>
    <w lemmaRef="#gloss_m73">mostrar</w> 
    las <w lemmaRef="#gloss_g16">gens</w> e los 
ostals.</item>
\end{minted}
\caption{État original des fichiers sources de \cite{lodge_les_2006}}
\label{fig:MontfOrig}
\end{figure}

\begin{figure}
\begin{minted}[linenos]{xml}
<entry n="116" xml:id="gloss_a116">
  <form>aver</form>
  <form source="#DOM" type="lmlv" cert="high">avẹr</form>
  <gramGrp>
    <pos>verbe</pos>
    <subc>transitif</subc>
    <mood>infinitif</mood>
  </gramGrp>
  <re xml:id="gloss_a116_1">
    <form>avem</form>
    <gramGrp>
      <pos>verbe</pos>
      <mood>indicatif</mood>
      <tns>présent</tns>
      <per>4</per>
    </gramGrp>
  </re>
<!-- Suite de l'entrée -->
</entry>
\end{minted}
\caption{Alignement des entrées du glossaire de \cite{lodge_les_2006} avec les lemmes du \textsc{dom} \parencite{stempel_dictionnaire_1996}}
\label{fig:alignLemmesMontf}
\end{figure}

Une fois l'alignement fait, nous avons réinjecté l'information de lemme dans les données segmentées (fig.~\ref{fig:MontfFinal}).

\begin{figure}
\begin{minted}[linenos]{xml}
<item xml:id="CC6.278">
<w lemma="item" xml:id="w_028267">Item</w>
<w lemma="@num@" xml:id="w_028268">IIII</w>
<w lemma="sol" xml:id="w_028269">s</w>
<pc>.</pc>
<w lemma="que" xml:id="w_028270">qu</w>
<w lemma="avẹr" xml:id="w_028271" lemmaRef="#gloss_a116_11">ac</w>
<w lemma="D" xml:id="w_028272">D</w>
<pc>.</pc>
<w lemma="Chapus" xml:id="w_028273">Chapus</w>
<w lemma="da+lo2" xml:id="w_028274" lemmaRef="#gloss_d57">dos</w>
<w lemma="@num@" xml:id="w_028275">II</w>
<w lemma="jọrn" xml:id="w_028276">jorns</w>
<w lemma="desus" xml:id="w_028277" lemmaRef="#gloss_d40">desus</w>
<w lemma="dire" xml:id="w_028278" lemmaRef="#gloss_d49_9">dit</w>
<w lemma="que" xml:id="w_028279" lemmaRef="#gloss_q11">que</w>
<w lemma="anar" xml:id="w_028280" lemmaRef="#gloss_a47_4">anet</w>
<w lemma="ẹn" xml:id="w_028281" lemmaRef="#gloss_e9">en</w>
<w lemma="Garsias" xml:id="w_028282">Garsias</w>
<w lemma="pẹr" xml:id="w_028283" lemmaRef="#gloss_p30">per</w>
<w lemma="mostrar" xml:id="w_028284" lemmaRef="#gloss_m73">mostrar</w>
<w lemma="lo2" xml:id="w_028285">las</w>
<w lemma="gẹn" xml:id="w_028286" lemmaRef="#gloss_g16">gens</w>
<w lemma="e" xml:id="w_028287">e</w>
<w lemma="lo2" xml:id="w_028288">los</w>
<w lemma="ostal" xml:id="w_028289">ostals</w>
<pc>.</pc>
</item>
\end{minted}
\caption{État final des fichiers des \textit{Comptes…}}
\label{fig:MontfFinal}
\end{figure}

Nous avons procédé d'une manière similaire pour le \textit{Flamenca} lemmatisé fourni par \cite{scrivner_old_2016}.
À l'issue de ce travail, une ultime phase de reprise d'ensemble, pour une correction et harmonisation générale, des données s'est avérée nécessaire. Pour ce faire, l'utilisation d'un tableur peut suffire, mais des logiciels plus perfectionnés existent, qui permettent la vérification de l'intégrité des lemmes et étiquettes utilisées, tout en facilitant la correction par lots des cas similaires \parencite[par ex.][]{clerice_pandora-postcorrect-app_2017}.

Nous disposions ainsi d'un premier (petit) corpus de 54600 occurrences, 
mais les lemmatiseurs de nouvelle génération, comme Pandora ou Lemming, fournissent aussi parfois la possibilité d'utiliser un corpus secondaire, non annoté, mais fournissant une grande quantité d'occurrences. Ce corpus secondaire peut être utilisé pour améliorer la modélisation de l'espace sémantique de la langue, en proposant une représentation des formes en fonction de leur contexte (à gauche et à droite) s'appuyant sur une quantité plus importante (et donc plus fiable) de données.
Pour cela, Pandora a recours à la technologie de \og{}\textit{word embeddings}\fg{} (ou, plongement de mots, en français) fournie par l'algorithme \og{}skipgram\fg{} de \texttt{w2vec} \parencite{mikolov2013distributed}. Lemming utilise l'algorithme \og{}Marlin\fg{} \parencite{martin1998algorithms}.

Pour ce corpus secondaire, nous avons extrait et utilisé les textes du premier volume des COM \parencite{ricketts_concordance_2001}, ce qui a nécessité de les convertir en \textsc{tei} et de les segmenter en mots.
	
\subsection{Entraînement d'un modèle de lemmatisation}
	
Nous avons retenu l'utilisation du lemmatiseur Pandora \parencite{kestemont_pandora:_2016}, sans pouvoir utiliser en parallèle \texttt{Lemming} \parencite{muller2015joint}, car, au cours des opérations d'alignement du corpus, nous n'avons pour l'instant traité que les lemmes et pas les étiquettes morpho-syntaxiques (pour lesquelles un jeu consensuel d'étiquettes n'existe pas à notre connaissance pour l'occitan ancien).
Lemming prenant en compte de pair morpho-syntaxe et lemmes, son utilisation devenait, pour le moment, impossible, tandis que Pandora les traite séparément. À l'avenir, il sera utile de compléter le corpus sur ce point pour permettre la comparaison des résultats de ces deux outils.

Pandora est un lemmatiseur en cours de développement, en Python, utilisant des bibliothèques d'intelligence artificielle telles que \texttt{Keras} et \texttt{TensorFlow}. Son code est libre et ouvert. Il a été conçu spécifiquement pour traiter les particularités des états médiévaux de langue, et notamment leur forte variabilité graphique.
Concrètement, Pandora utilise des plongements de mots et de caractères pour créer une représentation, par des vecteurs, des mots, de leurs caractères et de leur contexte à gauche et à droite. Cette représentation est ensuite fournie à un réseau de neurone convolutif ou récurrent (de type \textsc{lstm}) qui apprend sur les données du corpus et réalise la tâche de classification \parencites{kestemont_lemmatization_2016}{kestemont2017integrated}.
	
\begin{figure}
    \centering
    \includegraphics[width=\textwidth]{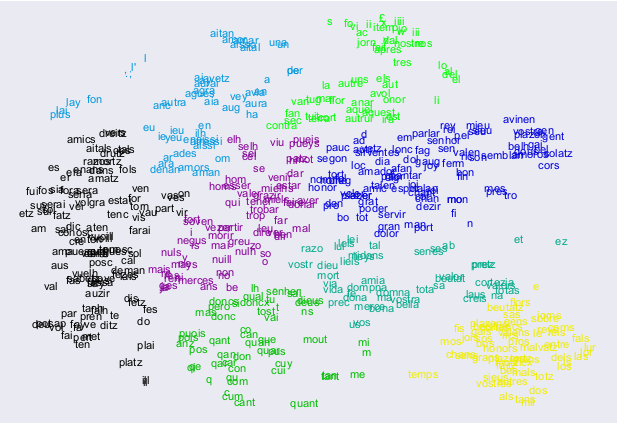}
    \caption{Visualisation en deux dimensions, par l'algorithme \textsc{t-sne}, des vecteurs de mots en 100 dimensions obtenus par l'application de l'algorithme \textsc{skipgram} sur les données des \textsc{com} et de notre corpus. Les mots sont colorés en fonction d'un partitionnement par classification ascendante hiérarchique (méthode de Ward, distance euclidienne).}
    \label{fig:embeddings}
\end{figure}

Nous avons ainsi pu entraîner des modèles de lemmatisation avec Pandora. Dans un premier temps, les données des \textsc{com} ont servi pour réaliser la représentation sémantique initiale des mots, en 100 dimensions avec l'algorithme \textsc{skipgram}. Cette représentation peut être visualisée par réduction en deux dimensions (fig.~\ref{fig:embeddings}), et donne à voir des proximités sémantiques. Celles-ci rapprochent des graphies différentes du même lemme, mais aussi des mots de sens voisin (i.e. employés dans des contextes similaires). Ces proximités peuvent aussi être interrogées directement et affichées avec un coefficient de similarité. Ainsi, les mots les plus proches de \textit{domna} sont \textit{dompna}  (0,97), \textit{dona} (0,87), \textit{bona} (0,86), \textit{amia} (0,85), etc. Ceux les plus proches de \textit{joy} sont \textit{gaug} (0,92), \textit{deport} (0,88), \textit{ioy} (0,85), \textit{esperansa} (0,85), \textit{pessamen} (0,84), etc.
	
L'entraînement peut ensuite commencer sur les données annotées. De celles-ci, 80\% sont utilisées pour l'entraînement, 10\% pour des ajustements, et 10\% pour tester la performance des modèles.
Durant chaque époque de l'entraînement, l'ensemble du corpus d'entraînement est soumis au réseau de neurones convolutif, qui apprend et tente de prédire les lemmes. Après un nombre donné d'époques (ici, 100), l'entraînement est arrêté. Les résultats obtenus par le meilleur modèle sont présentés en table~\ref{tab:pandoraResults}. Si la performance sur les formes connues est bonne, elle demeure très nettement en deçà de ce qu'il est possible d'atteindre sur les formes inconnues avec une quantité supérieure de données d'entraînement.
	
\begin{table}
    \centering
    \begin{tabular}{c|ccc}
         & Entr. & Dev. & Test \\ \hline \hline
toutes formes & 0,99 & 0,86 & 0,88 \\
formes connues & 0,99 & 0,92 & 0,93 \\
formes inconnues & NA & 0,35 & 0,37
    \end{tabular}
    \caption{Précision des résultats obtenus par le modèle entraîné avec Pandora après 100 époques d'apprentissage, sur les données d'entraînement, d'ajustement et de test}
    \label{tab:pandoraResults}
\end{table}

	\section{Des perspectives d'exploitation renouvelées}
	
	Si les traitements présentés ici sont d'ores-et-déjà fonctionnels, et permettent d'obtenir des résultats intéressants en production de corpus, la clef de leur amélioration pour l'avenir consiste en le nécessaire accroissement de la quantité de données disponibles et utilisables. Une quantité supérieure de données d'entraînement signifie en effet l'obtention de modèles plus performants qui permettent en retour un accroissement plus rapide de la quantité de données exploitables.
	
	Pour cela, il faut que les données soient interopérables et documentées, mais aussi rendues disponibles en ligne, sous une licence libre, permettant la réutilisation, et sous une forme citable et pérenne. Ce n'est qu'à ce prix que l'on permettra un progrès des outils, méthodes et connaissances sur les textes occitans médiévaux et pré-modernes.

	\section*{Remerciements}
	Les auteurs remercient Marine Mazars et Lucence Ing pour leur collaboration philologique et technique, ainsi que Thomas Fields et Olga Scrivner pour la communication de leurs données. Cet article s'inscrit dans une démarche relevant des projets LAKME (financement de l'Université Paris Sciences \& Lettres) et CoRLiG (ministère de la culture, DGLFLF).

	\renewcommand*{\mkbibnamefamily}[1]{\textsc{#1}}
	\printbibliography
	
\end{document}